\documentclass[sigconf]{acmart}

\usepackage{algorithm}
\usepackage{algorithmic}
\usepackage{graphicx}
\usepackage{amsmath}
\usepackage{breqn}
\usepackage{listings}

\AtBeginDocument{%
  }

\copyrightyear{2025}
\acmYear{2025}
\setcopyright{acmlicensed}\acmConference[CIKM '25]{Proceedings of the 34th
ACM International Conference on Information and Knowledge
Management}{November 10--14, 2025}{Seoul, Republic of Korea}
\acmBooktitle{Proceedings of the 34th ACM International Conference on
Information and Knowledge Management (CIKM '25), November 10--14, 2025,
Seoul, Republic of Korea}
\acmDOI{10.1145/3746252.3761117}
\acmISBN{979-8-4007-2040-6/2025/11}

\begin{document}

\title[Generating Intent-Driven Dialogues with Contrastive Learning for Multi-Turn Classification]{From Intents to Conversations: Generating Intent-Driven Dialogues with Contrastive Learning for Multi-Turn Classification}


\author{Junhua Liu}
\authornote{Equal contributions.}
\affiliation{%
  \institution{Forth AI}
  \country{Singapore}
}
\email{j@forth.ai}


\author{Yong Keat Tan}
\authornotemark[1]
\affiliation{%
  \institution{Shopee}
  \country{Singapore}
}
\email{yongkeat.tan@shopee.com}

\author{Bin Fu}
\authornote{Corresponding author.}
\affiliation{%
  \institution{Shopee}
  \country{Singapore}}
\email{bin.fu@shopee.com}

\author{Kwan Hui Lim}
\affiliation{%
  \institution{Singapore University of Technology and Design}
  \country{Singapore}}
\email{kwanhui@acm.org}

\renewcommand{\shortauthors}{Liu et al.}

\begin{abstract}
In conversational AI systems, a critical challenge in training effective multi-turn intent classification models lies in the generation of large-scale, domain-specific, multilingual dialogue datasets. In this paper, we introduce Chain-of-Intent, a novel framework that integrates Hidden Markov Models (HMMs) with Large Language Models (LLMs) to generate intent-driven, context-aware dialogues through self-play. Our method first extracts domain-specific intent transition patterns from real-world e-commerce chat logs, which guide the modeling of turn-level dynamics and intent sequences. LLMs are then employed to parameterize the emission probabilities of HMMs, enabling the generation of natural, coherent utterances aligned with predicted intents and dialogue context. We also propose MINT-CL, a multi-task contrastive learning framework for multi-turn intent classification, which improves performance while reducing dependence on large-scale annotated datasets. Empirical results demonstrate that our approach outperforms competitive baselines in dialogue generation quality and classification accuracy, particularly in multilingual settings. To facilitate future research, we release MINT-E, a comprehensive, multilingual, intent-aware multi-turn dialogue corpus derived from the e-commerce domain\footnote{The reproduced source code and dataset are available at \url{https://github.com/junhua/chain-of-intent}.}.
\end{abstract}

\ccsdesc[500]{Information systems~Language models}
\ccsdesc[500]{Computing methodologies~Intelligent agents}
\ccsdesc[500]{Information systems~Question answering}

\keywords{Multi-turn Intent Classification, Large Language Model, Multi-task Learning, Contrastive Learning}
\begin{teaserfigure}
  \centering
  \vspace{-3mm}
  \includegraphics[width=0.85\textwidth]{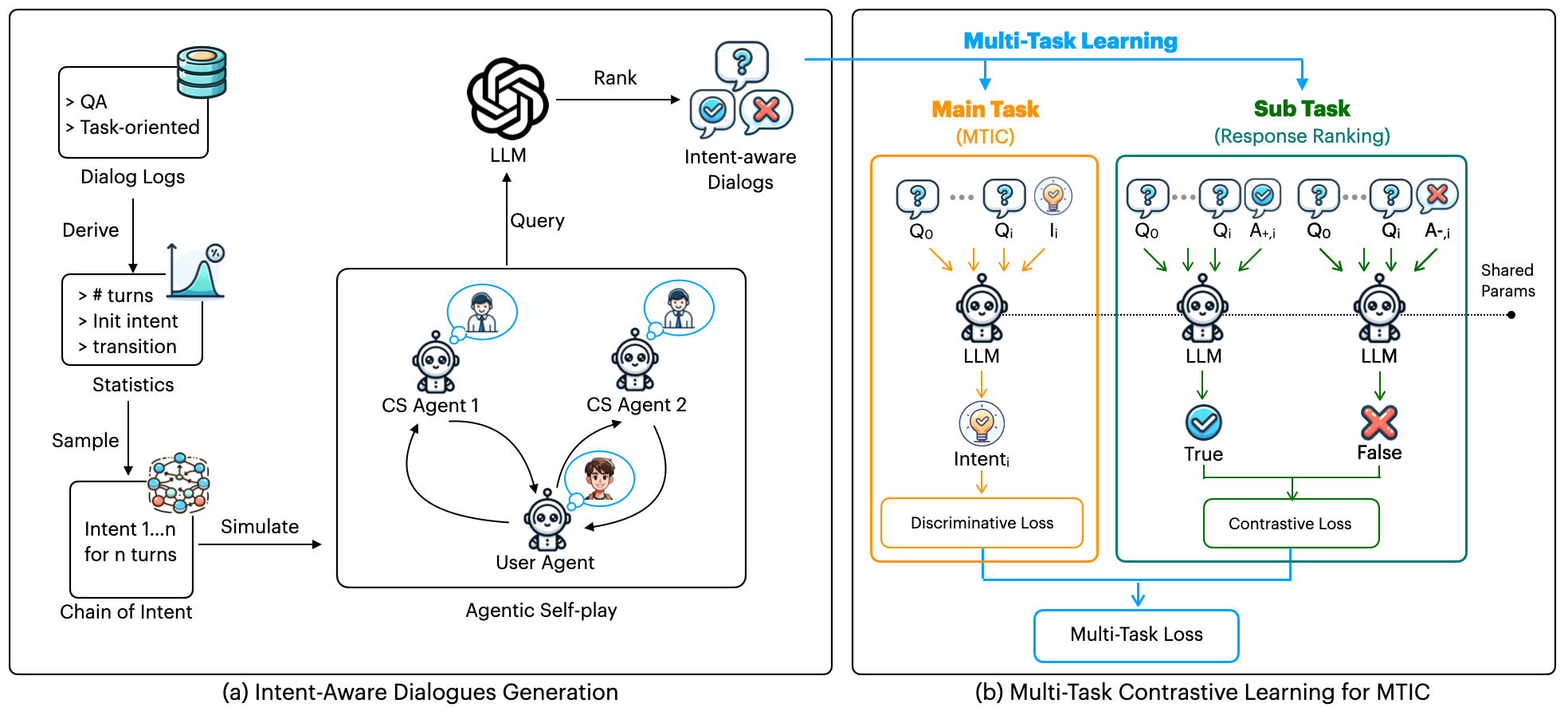}
  \vspace{-3mm}
  \caption{Overview of the proposed pipeline to sample a chain-of-intent and the subsequent intent-aware dialogues generation (a), and the multi-task contrastive learning model for multi-turn intent classification downstream task (b).}
  \label{fig:teaser}
\end{teaserfigure}


\maketitle

\section{Introduction}

Chatbots are essential on international e-commerce platforms, offering 24/7 customer service. Users typically interact with these chatbots through multi-turn conversations to achieve their goals. Therefore, it is crucial for the chatbot to accurately identify the user's intent at each turn using a Multi-Turn Intent Classification (MTIC) model. To improve accuracy, MTIC models should consider not just the current message but also previous interactions and intents. Ignoring the context from earlier turns often leads to mistakes, giving irrelevant answers and making users dissatisfied~\cite{xu2014contextual}.

\textbf{Motivation and Challenges}. Like many deep learning approaches~\cite{Soudani2024ASO}, MTIC models heavily rely on large amounts of domain-specific multi-turn intent-utterance pairs to boost their performance~\cite{ni2023recent}. However, collecting such datasets efficiently, especially across different languages, is challenging. While some studies on dialogue understanding in multi-turn intent classification~\cite{wu2021context,qu2019user,ren2020intention} assume there is plenty of multi-turn training data available, that is often not the case in reality. Recent work~\cite{zhao2023chatgpt} shows that using few-shot prompting with ChatGPT does not perform well for intent classification, and it does not match up to supervised methods.

\begin{figure}[t]
    \centering
    \includegraphics[width=.45\textwidth]{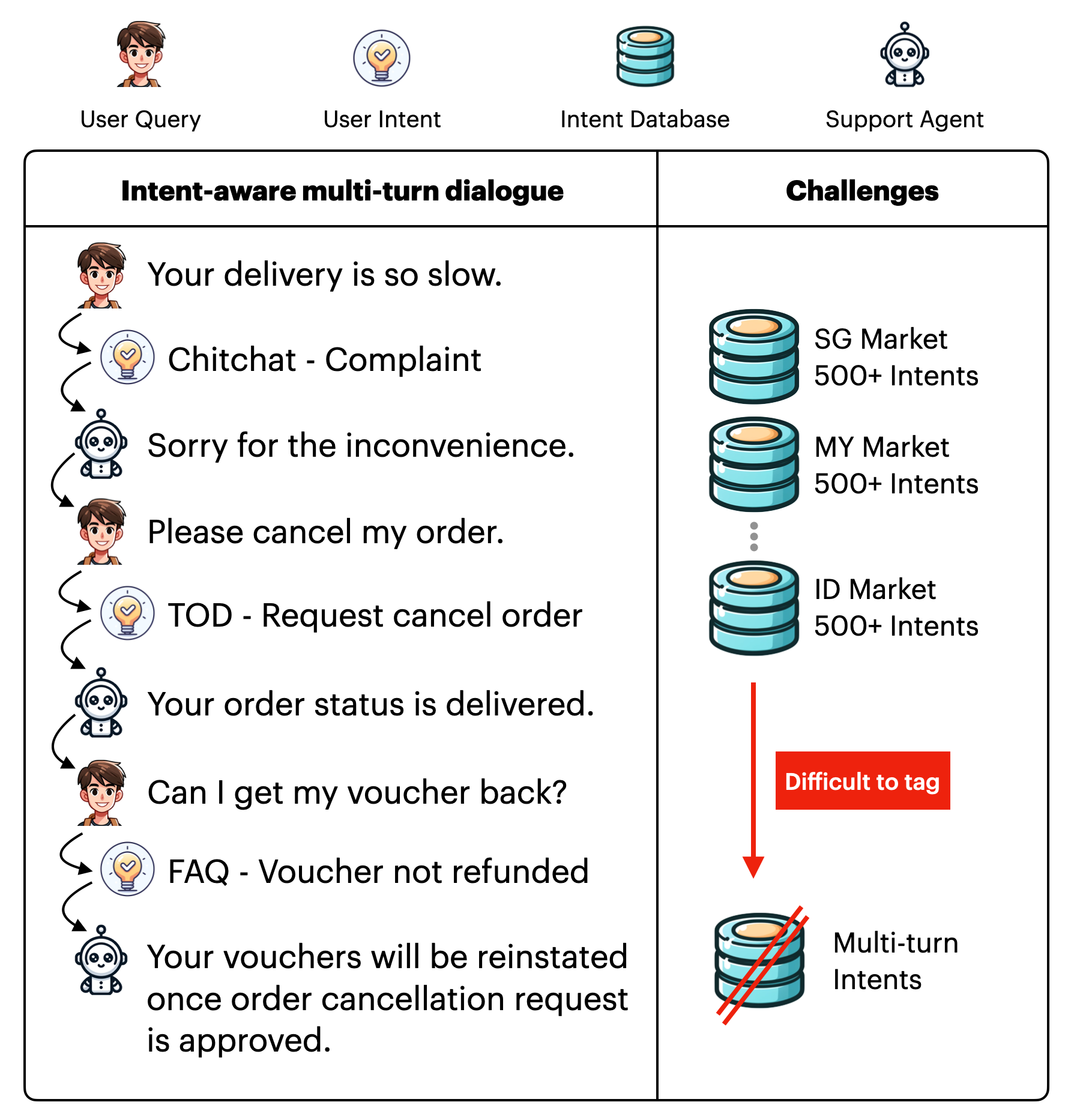}
    \vspace{-0.4cm}
    \caption{Example of intent-aware multi-turn dialogue and the challenges of annotating single-turn cross-market intents to create a multi-turn intent classification dataset. The intents (e.g., Chitchat - Complaint) are in the format of <intent type>-<intent name>.}
    \label{fig:challenge}
    \vspace{-.6cm}
\end{figure}

Figure~\ref{fig:challenge} illustrates the challenges in annotating multi-turn dialogues. For MTIC in information-seeking dialogues~\cite{Askari2024SelfseedingAM}, manual annotation of intents is time-consuming and resource-intensive, which leads to a lack of training data. Unlike popular intent classification datasets such as MSdialog-Intent~\cite{Qu2018AnalyzingAC}, which has only 12 unique intents, industrial chatbots handle hundreds of intents to meet specific user needs in each market. This vast number of intents significantly increases the complexity of both classification tasks and data annotation. With so many options, annotators can easily make errors and need more time to make the right decisions, making the process expensive and slow. For instance, \citet{qu2019user} reported that labelling intents for 2,199 dialogues cost \$1,700. Manually annotating large-scale, multi-turn datasets in multiple languages or constantly updating the model for new user intents is impractical. Lack of sufficient training data often leads to poorer performance, highlighting the need for a better way to address this challenge. 

Most research integrates the dialogue acts (DA) for domain-specific task-oriented dialogue (TOD) generation \cite{Wang2020MultiDomainDA}. However, dialogue acts are more general, like statement/request/yes-answer, and therefore unusable for domain-specific generation. Recent works also adopt specific DA in the e-commerce domain for only TOD intents \cite{Chen2019TheJC}. However, there are FAQ, chit-chat, and TOD intents in multi-turn sessions shown in Figure ~\ref{fig:challenge}, so considering only TOD intents is limited for dialogue generation in real-world scenarios.

To address these issues, this paper contributes three solutions from different aspects, as follows:

\textbf{Contribution 1: Chain-of-Intent}. We introduce \textbf{Chain-of-Intent}, a novel method for generating intent-aware dialogues through self-play. Our approach combines Hidden Markov Models (HMMs) to model intent sequences with Large Language Models (LLMs) to produce coherent, context-aware multi-turn dialogues. By extracting domain-specific knowledge from historical chat logs, we generate realistic intent sequences and leverage LLMs to create high-quality multilingual dialogues. This method addresses the data scarcity problem by generating large-scale, domain-specific dialogue datasets without extensive manual annotation.




\textbf{Contribution 2: MINT-CL}. We propose \textbf{MINT-CL}, a framework for MTIC with multi-task contrastive learning. MINT-CL enhances the accuracy of MTIC models by leveraging both current and contextual information and optimizing the model with a multi-task contrastive learning objective. Our experiments demonstrate that MINT-CL outperforms baseline models across various languages and markets.

\textbf{Contribution 3: MINT-E}. Finally, we release \textbf{MINT-E}, a multilingual intent-aware multi-turn e-commerce dialogues corpus, to support research in this field. The corpus covers eight different markets with diverse languages and a large number of intents (as shown in Table~\ref{table:data_train_test}), providing a valuable resource for developing and evaluating MTIC models. 




\section{Intent-aware Question Answer Dialogues}
\subsection{Single-turn} 

In a closed domain question answering (QA) system, we have a set of predetermined intent classes $\mathcal{I} = \{I_i\}_{i=1}^k$, ranging from informational queries to action-oriented requests. Here, we assume that a dataset of single-turn question-intent pairs $\mathcal{D} = {(x_i, y_i)} $ can be easily accessible, where $x_i$ denotes the question and $y_i \in \mathcal{I}$ is its labelled intent. 
This dataset is described as single-turn because all $x_i$ on their own are standalone complete utterances and are semantically clear for intent inference.

\subsection{Multi-turn} 
The key difference between single-turn and multi-turn QA is the presence of history. In a multi-turn conversation $\mathcal{C}$ with number of turns $t$, we have an alternating sequence of $t$ question and answer pairs, and each question is associated with an intent. Formally, $\mathcal{C} = \{(q_1, I_1, a_1),..., (q_{t}, I_{t}, a_{t})\}$, where $q_t$ is the question at $t$-th turn, $I_t$ is its intent, and $a_t$ is the corresponding answer. We now define the task for e-commerce multi-turn conversation question generation. Given an intent $I_t$, and a conversation history $h_t = \{(q_1, a_1),..., (q_{t-1}, a_{t-1})\}$, the multi-turn conversation question generation task aims to generate a question $q_t$ at $t$-th turn:
$$ q_t = \arg \max_q P(q | I_t, h_t) $$
in which the generated question should be coherent with the conversation history and natural like in human conversation.


\begin{figure}[t]
    \centering
    \includegraphics[width=.48\textwidth]{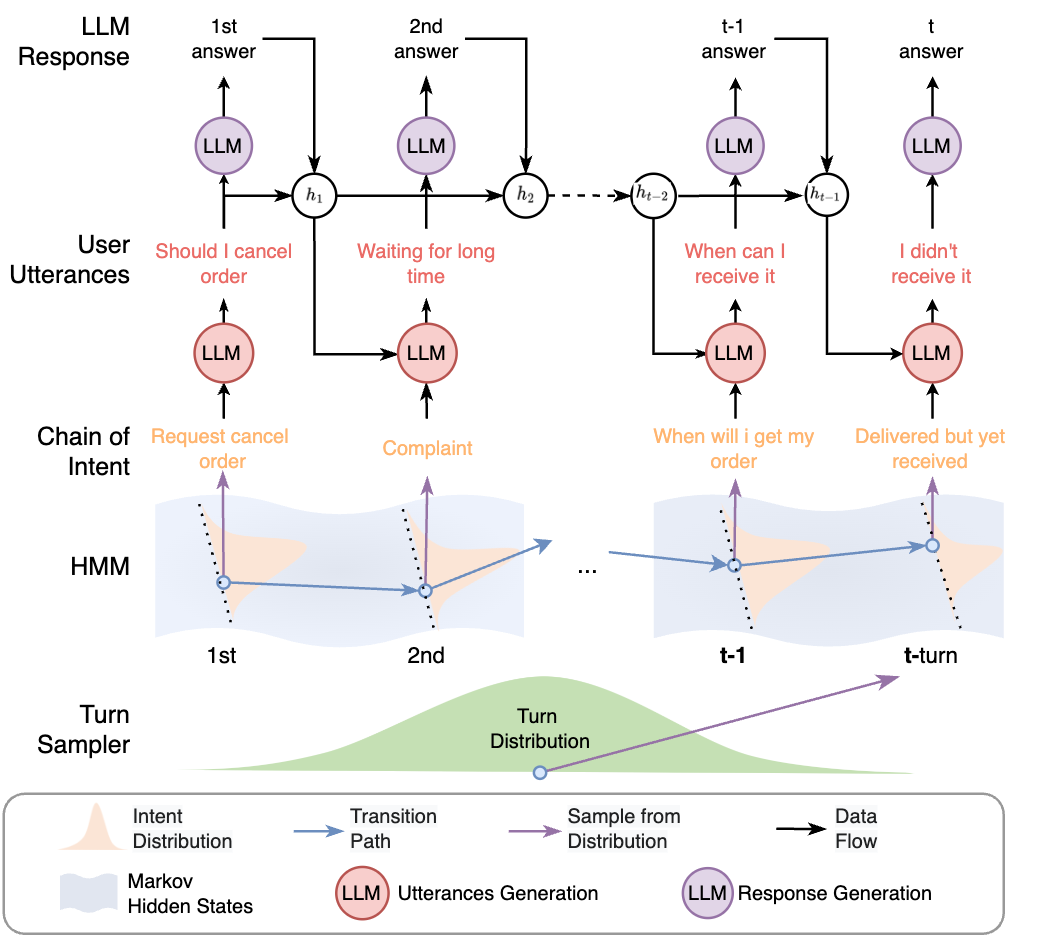}
    \vspace{-.6cm}
    \caption{Overview of the Chain-of-Intent framework: using LLM-enhanced HMM to sample intent sequences and generate intent-aware multi-turn dialogues.}
    \label{fig:method}
    \vspace{-0.3cm}
\end{figure}

\section{Intent-aware Dialogue Generation}
Figure~\ref{fig:method} shows the detailed workflow of intent-aware dialogue generation, which comprises two main stages: domain knowledge extraction and question generation.

\subsection{Domain Knowledge Extraction}
As LLMs do not possess domain-specific knowledge out-of-the-box, this step is imperative to extract the knowledge to be injected into the conversation generation process, bridging the gap for business usage. There are two types of domain-specific knowledge to account for when generating a conversation, namely the number of turns in each conversation and the intent transition from intents of historical questions. These two combined will guide the generation of each conversation. 

\subsubsection{Turn Number Estimation}
From the real-life chat logs between the platform user and agent, the probability distribution of the number of turns of each conversation $P(T)$ is observed. Before generating a conversation, the number of turns in the conversation $T$ is first decided by sampling from $P(T)$.

\subsubsection{Intent Transition Matrix}
Other than the number of turns, we also need to know the sensible sequence of intents that users might have in a conversation. Just like the number of turns, the intent transition information is also gathered from the chat logs using Laplace smoothing with $\alpha = 0.1$ to handle unseen transitions. One limitation here is that the intent from chat logs is inferred by an intent recognition model trained on the single-turn data $\mathcal{D}$. Each $q_t$ is inferred separately without considering $h_t$ to obtain $I_t$ for the $t$-th turn. That said, by the Law of Large Numbers, if we gather the statistics from a reasonably large number of chat logs, the intent transition matrix obtained will provide a good estimation of true intents. Hereby, the initial intent distribution $P(I_1)$ is specified in a row matrix, while the intent transition $P(I_t|I_{t-1})$ is defined in another $|\mathcal{I}| \times |\mathcal{I}|$ matrix. Note that the $P(I_t|I_{t-1})$ here is time invariant, the probability of $P(I_j|I_i)$ will be the same regardless of $t$. 

Using this statistical approach, we are able to sample a probabilistic chain of intent with the sampled number of turns and intent transition matrix. The sampled chain of intent plays a crucial role in the following question generation process.

\subsection{Chain-of-Intent}
\label{ori_hmm}

We introduce \textbf{Chain-of-Intent}, an effective method for self-play generation of intent-aware dialogues. Our approach utilizes a HMM to generate sequences of user intents in specific domains and integrates LLMs to produce coherent and context-aware dialogues.

\subsubsection{Intent Sampling with HMM}

The HMM models the user's interaction with the chatbot, where the intents are the hidden states and the utterances are the observations. To capture realistic intent flows within specific domains, we extract domain knowledge from historical chat logs by calculating:

\begin{itemize}
    \item The turn distribution $P(T)$: the probability distribution of the number of turns in a conversation.
    \item The initial intent distribution $P_{\text{init}}$: the probability distribution of intents at the first turn.
    \item The intent transition distribution $P_{\text{trans}}$: the probability of transitioning from one intent to another.
\end{itemize}

Using these distributions, we first sample the number of turns $T$ from $P(T)$, then sample the initial intent $I_1$ from $P_{\text{init}}$, and subsequent intents $I_t$ from $P_{\text{trans}}(\cdot|I_{t-1})$ conditioned on the previous intent.

Formally, we define the HMM as:
\vspace{-.1cm}
\begin{equation}
P(I_{1:T}, q_{1:T}) = P(I_1) P(q_1|I_1) \prod_{t=2}^{T} P(I_t|I_{t-1}) P(q_t|I_t)
\end{equation}

where $I_t$ corresponds to the intent at turn $t$, and $q_t$ corresponds to the user's utterance (question) at turn $t$, with normalization constraints $\sum P(T) = 1$ and $\sum P(I_t|I_{t-1}) = 1$. The observation model $P(q_t|I_t)$ is modeled by uniformly sampling a question $X_t$ from the set of questions in the single-turn dataset $\mathcal{D}$ that are labeled with intent $I_t$:
\vspace{-.1cm}
\begin{equation}
P(X_{1:T}, I_{1:T}) = P(I_1) P(X_1|I_1) \prod_{t=2}^{T} P(I_t|I_{t-1}) P(X_t|I_t)
\label{eq:hmm_qi}
\end{equation}

After sampling intents and utterances according to this model, we obtain a sequence of questions $q_{1:T}$.

\subsubsection{Limitations of Classic HMMs}

While HMMs can model sequences of intents and generate corresponding utterances, they have significant limitations in capturing the dependencies present in natural human conversations. Specifically, classic HMMs assume that observations (utterances) are conditionally independent given the current state (intent), and that the next state depends only on the current state. This independence assumption neglects the rich context and dependencies in multi-turn dialogues, such as anaphora resolution and ellipsis. For example:

\begin{quote}
    \textbf{Person A}: "Did you see Mary yesterday?"\\
    \textbf{Person B}: "Yes, she was at the park." (Anaphora resolution: "she" refers to Mary)\\

    \textbf{Person A}: "Did she say anything interesting?"\\
    \textbf{Person B}: "Not much, just talked about her new job."\\
    \textbf{Person A}: "And John?"\\
    \textbf{Person B}: "Didn't see him." (Ellipsis: "I" is omitted from full response "I didn't see him".)
\end{quote}

In contrast, questions generated by the HMM tend to be standalone and overly explicit, often repeating context unnecessarily and lacking coherence between turns. Questions from different turns may also semantically or logically contradict each other due to the independence assumptions, resulting in unnatural conversation flows.

\subsubsection{Enhancing HMMs with LLMs}

To overcome these limitations, we integrate LLMs into the HMM framework, allowing us to generate dialogues that are coherent and context-aware. Specifically, we leverage the LLM's in-context learning ability to generate each user utterance $q_t$ and the corresponding chatbot response $a_t$, conditioning on the conversation history up to turn $t-1$.

For question generation at turn $t$, we construct an instruction prompt $\mathcal{S}_q(I_t, X_t)$ that includes the current intent $I_t$ and a sample question $X_t$ associated with $I_t$ (to provide context and clarify the intent). This prompt guides the LLM to generate a user question $q_t$ that is coherent with the conversation history $h_t = \{(q_1, a_1), \dots, (q_{t-1}, a_{t-1})\}$. The complete input prompt for question generation is:
\vspace{-.05cm}
\[
\mathcal{P}_q = \{\mathcal{S}_q(I_t, X_t), h_t\}
\]

Similarly, for answer generation, we provide a simple guidance $\mathcal{S}_a$ along with the conversation history and the current question $q_t$. The complete input prompt for answer generation is:
\vspace{-.05cm}
\[
\mathcal{P}_a = \{\mathcal{S}_a, h_t, q_t\}
\]

By generating both questions and answers in this way, the LLM can utilize all previous utterances and intents, effectively capturing dependencies and producing more natural dialogues.

To incorporate contextual dependencies, we extend the formulation in Equation~\ref{eq:hmm_qi} by conditioning question generation on conversation history:
\vspace{-.05cm}
\begin{align}
P(q_{1:T}, I_{1:T}) = &\, P(I_1) P(X_1 | I_1) P(q_1 | I_1, X_1) \notag \\
& \times \prod_{t=2}^{T} P(I_t | I_{t-1}) P(X_t | I_t) P(a_{t-1} | q_{1:t-1}) \notag \\
& \times P(q_t | I_t, X_t, q_{1:t-1}, a_{1:t-1})
\label{eq:hmm_qi_sub}
\end{align}

This formulation captures the dependencies between the generated questions and answers and their history, overcoming the limitations of classic HMMs. 

This generative process results in a conversation:
\[
C = \{(q_1, I_1, a_1), \dots, (q_T, I_T, a_T)\}
\]

\subsubsection{Answer Ranking}
\label{sec:answer_ranking}

For each conversation, we generate an alternative response for the last turn using a smaller LLM, namely \textit{Meta-Llama-3-8B-Instruct}~\cite{dubey2024llama}. This produces an alternative conversation $C_{\text{llama}}$, where the last answer is replaced with $a_T^{\text{llama}}$.

To evaluate and compare the quality of the responses, we perform a point-wise evaluation using GPT-4, asking it to rate the quality of the last answer in each conversation on a scale from 1 to 10, given the chat history and the last question, we use an input prompt:
\vspace{-.05cm}
\[
\mathcal{P}_e = \{\mathcal{S}_e, h_T, q_T, a_T\}
\]

where $\mathcal{S}_e$ is the instruction for evaluation.

By comparing the scores of $C$ and $C_{\text{llama}}$, we can determine which answer is better. This ranking information is useful for downstream tasks, such as training the MTIC model with multi-task contrastive learning, as described in Section~\ref{sec:mtic}.

\subsection{MINT-E}
Using the proposed pipeline, we present \textbf{MINT-E}, a large multilingual intent-aware multi-turn e-commerce dialogues corpus. The statistics of MINT-E, together with two existing dialogue corpus, namely Ubuntu Dialogue Corpus (UDC) and MSDialog, is shown in Table \ref{table:statistics}. The statistics of MINT-E are obtained by averaging across all markets. Only markets with language separable by white spaces, such as BR, ID, MY, PH, SG, and VN, are included.

As ID is the main market with the most traffic online, we observe the top 15 intents based on the question-level ratio in ID generated dataset. The top user intent is to check delivery status, which makes up to 13.8\% of the questions. We also observed a long-tailed distribution with the top 10 intents taking up more than 50\% of the questions.

Notably, while MSDialog contains only 12 unique intents and UDC lacks intent annotations altogether, MINT-E encompasses 381 distinct intents. This extensive range allows for more granular intent classification, reflecting the complexity and variety of real-world user interactions in e-commerce settings.

Moreover, MINT-E is a multilingual corpus, covering eight different markets with diverse languages. In contrast, both UDC and MSDialog are predominantly in English. This multilingual aspect of MINT-E makes it particularly valuable for developing and evaluating MTIC models in diverse linguistic environments, addressing the challenges associated with intent classification across different languages and cultures.

Although MINT-E has fewer sessions and questions compared to UDC and MSDialog, the data is richer in terms of intent diversity and domain specificity. The average number of words per question in MINT-E is around 11, similar to UDC, which suggests concise and natural conversational exchanges typical in chatbot interactions. MSDialog, on the other hand, has a much higher average of 75.91 words per question, indicating longer and more complex queries that may not reflect typical user behavior in e-commerce platforms.

Additionally, MINT-E focuses specifically on e-commerce dialogues, providing domain-specific context that is highly relevant for training and evaluating chatbot systems in this sector. The corpus includes real-world scenarios and user intents that are common in online shopping experiences, making it a practical resource for both researchers and practitioners.

In summary, MINT-E's advantages over benchmark datasets like UDC and MSDialog include:
\begin{enumerate}
    \item Extensive Intent Coverage: With 381 unique intents, MINT-E allows for detailed and nuanced intent classification, capturing the complexities of user interactions in e-commerce.
    \item Multilingual Support: Covering eight markets and languages, MINT-E addresses the need for multilingual MTIC models, facilitating research in global contexts.
    \item Domain Specificity: Focused on e-commerce, the dataset provides relevant and practical dialogues that are directly applicable to industry needs.
    \item Realistic Conversational Data: The average length of questions and the number of turns per session reflect actual user behavior, enhancing the applicability of models trained on this data.
\end{enumerate}

As a rich, diverse, and practical dataset, MINT-E stands as a valuable resource that overcomes the limitations of existing corpora, supporting the development of more accurate and context-aware chatbot systems in multilingual and domain-specific applications.

\begin{table}[t]
\centering
\scalebox{.95}{
\begin{tabular}{|l| c c c|} 
 \hline
 Items & UDC & MSDialog & MINT-E \\ [0.5ex] 
 \hline
 \# Intents & - & 12 & 381 \\
 \# Sessions & 930k & 35k & 20k \\
 \# Questions & 7.1M & 300k & 44k \\
 \# Words & 100M & 24M & 482k \\
 \# Avg. \# questions per session & 7.71 & 8.94 & 2.92 \\
 \# Avg. \# words per question & 10.34 & 75.91 & 11.00 \\ [1ex]
 \hline
\end{tabular}}
\caption{Comparison of dataset statistics for UDC, MSDialog, and MINT-E.}
\label{table:statistics}
\vspace{-1cm}
\end{table}
\section{Multi-Turn Intent Classification}
\label{sec:mtic}

To assess the effectiveness of the MINT-E corpus, we focus on a relevant downstream task: Multi-Turn Intent Classification (MTIC). This section details the MTIC task, introduces our proposed training framework \textbf{MINT-CL}, which incorporates multi-task contrastive learning, and describes the experimental settings used to evaluate our approach.

\subsection{Task Definition}

In MTIC, the goal is to predict the user's intent at the current turn of a conversation by considering both the current utterance and the preceding dialogue context. Given a conversation with a sequence of user utterances $\{q_1, q_2, \dots, q_t\}$, the task is to determine the intent $I_t$ associated with the latest utterance $q_t$, utilizing the information from all previous turns. The input of the model concatenates all user utterances into a single sequence $q_{\text{all}}$, separated by commas:
\vspace{-.12cm}
\[
q_{\text{all}} = ``<q_1>, <q_2>, \dots, <q_t>"
\]

\subsection{Language Model}

The language model used for classification is designed to handle the hierarchical nature of intents, organized into a three-level taxonomy. To capture both local and global hierarchical information, we combine a label-attention mechanism with a hierarchical text classification (HTC) approach inspired by HiTIN~\cite{Zhu2023HiTINHT}.

\subsubsection{Text Representation}

We utilize the XLM-RoBERTa-base model to obtain contextual embeddings of the input text. Specifically, we extract the embedding corresponding to the [CLS] token to represent the entire input sequence:
\vspace{-.1cm}
\[
H = \Phi_{\text{XLM-R}}(q_{\text{all}})^{[\text{CLS}]} \in \mathbb{R}^{d}
\]

where $\Phi_{\text{XLM-R}}$ denotes the XLM-RoBERTa encoder, and $d$ is the dimension of the hidden state. The encoder is further pre-trained on our in-domain corpus to enhance its representation capabilities.

\subsubsection{Label-Attention Mechanism}

To leverage local information at each level of the intent hierarchy, we employ a label-attention mechanism with separate classifier heads for each layer. For each layer $l$, we compute an intermediate representation $L_l$:
\vspace{-.1cm}
\[
L_l = 
\begin{cases}
H W^1_l + b^1_l, & \text{if } l = 1, \\
(H \oplus L_{l-1}) W^1_l + b^1_l, & \text{if } l > 1,
\end{cases}
\]

where $W^1_l \in \mathbb{R}^{d \times d}$ for $l = 1$ and $W^1_l \in \mathbb{R}^{2d \times d}$ for $l > 1$, $b^1_l \in \mathbb{R}^{d}$, and $\oplus$ denotes concatenation. This setup allows each layer to incorporate information from the previous layer's output.

The logits for each layer are then calculated as:
\vspace{-.15cm}
\[
H_{\text{local}}^l = L_l W^2_l + b^2_l
\]

where $W^2_l \in \mathbb{R}^{d \times |\mathcal{I}_l|}$ and $b^2_l \in \mathbb{R}^{|\mathcal{I}_l|}$, with $|\mathcal{I}_l|$ being the number of classes at layer $l$.

\subsubsection{Hierarchical Text Classification Component}

To capture global hierarchical information, we integrate an HTC component inspired by HiTIN. In this approach, a tree network is constructed based on the intent taxonomy, and messages are propagated in a bottom-up manner. The text representation $H$ is broadcast to the leaf nodes, and through hierarchical message passing, we obtain global logits $H_{\text{global}}^l$ for each layer.

\subsubsection{Ensembling and Prediction}

We combine the local and global logits for each layer:
\vspace{-.15cm}
\[
Y_l = \text{softmax}\left( H_{\text{local}}^l + H_{\text{global}}^l \right)
\]

To ensure consistency across the hierarchy, we perform a beam search over the three layers to predict the final intent.

\subsection{MINT-CL}

To enhance the model's performance, we introduce \textbf{MINT-CL}, a multi-task learning framework that incorporates a response ranking task using contrastive learning for MTIC. This framework effectively aids LLM in learning more robust representations by distinguishing between high-quality and lower-quality responses.

\subsubsection{Response Ranking Task}

For each conversation, we generate two responses for the last turn: a high-quality response $a^+$ and a lower-quality response $a^-$. These responses are obtained from different models or configurations, such as using different language models or varying the generation settings.

We create input pairs by concatenating the conversation history with each response, separated by the [SEP] token:
\vspace{-.1cm}
\begin{align}
c^+ = q_{\text{all}} \; [\text{SEP}] \; a^+ 
 \quad
c^- = q_{\text{all}} \; [\text{SEP}] \; a^-
\end{align}

These inputs are then encoded using the same encoder:
\vspace{-.12cm}
\begin{align}
H_2^+ = \Phi_{\text{XLM-R}}(c^+)^{[\text{CLS}]}
\quad 
H_2^- = \Phi_{\text{XLM-R}}(c^-)^{[\text{CLS}]}
\end{align}

\subsubsection{Contrastive Loss}

We employ a contrastive loss similar to \cite{Gao2020DialogueRR} to encourage the model to assign higher scores to better responses:
\vspace{-.15cm}
\[
\mathcal{L}_{\text{contrastive}} = -\log \frac{\exp(L_2^+)}{\exp(L_2^+) + \exp(L_2^-)}
\]

where
\vspace{-.1cm}
\begin{align}
L_2^+ = H_2^+ W_2 + b_2
\quad
L_2^- = H_2^- W_2 + b_2
\end{align}

and $W_2 \in \mathbb{R}^{d \times 1}$ and $b_2 \in \mathbb{R}$ are the weights and bias for the ranking task.

\subsubsection{Combined Training Objective}

The final loss function is a weighted sum of the intent classification loss $\mathcal{L}_{\text{intent}}$ and the contrastive loss $\mathcal{L}_{\text{contrastive}}$:
\vspace{-.1cm}
\[
\mathcal{L} = \mathcal{L}_{\text{intent}} + \lambda \mathcal{L}_{\text{contrastive}}
\]

Here, $\lambda$ is a hyperparameter that balances the contribution of the contrastive loss. Based on validation performance across three values $\{0.1, 0.3, 0.5\}$, we set $\lambda = 0.3$.

\section{Experiments}
\label{sec:exp}

To evaluate our proposed methods, we conducted a series of experiments across datasets from eight different markets.  The assessment employed two main metrics: the quality of the generated dialogues and the accuracy on the downstream MTIC tasks. This section outlines the datasets used, the experimental setups, the experimental results and discussion.

\begin{table}[t]
\centering
\scalebox{1}{
\begin{tabular}{|c|c|c|c|}
\hline
Market & Language & \# Intents & \# Samples \\
\hline
BR & Portuguese (pt) & 312 & 67k \\
ID & Indonesian (id) & 503 & 162k \\
MY & English/Malay (en, ms) & 473 & 72k \\
PH & English/Filipino (en, fil) & 252 & 27k \\
SG & English (en) & 347 & 63k \\
TH & Thai (th) & 377 & 49k \\
TW & Traditional Chinese (zh-tw) & 387 & 32k \\
VN & Vietnamese (vi) & 399 & 188k \\
\hline
\end{tabular}}
\caption{Dataset statistics for each market, including primary languages, number of intents, and dataset sizes.}
\label{table:data_train_test}
\vspace{-.8cm}
\end{table}

\subsection{Datasets}
\label{sec:dataset}

We utilized e-commerce question-intent datasets from eight markets: Brazil (BR), Indonesia (ID), Malaysia (MY), the Philippines (PH), Singapore (SG), Thailand (TH), Taiwan (TW), and Vietnam (VN). Table~\ref{table:data_train_test} provides details on the primary languages, number of intents, and the size of the datasets for each market. Among these, SG, MY, and PH primarily use English, while the others communicate in their native languages during conversations.

To extract domain-specific knowledge, we collected about 100,000 online chatbot sessions from each market. This data helped us estimate the market-specific turn distribution $P(T)$, initial intent distribution $P(I_1)$, and intent transition distribution $P(I_t | I_{t-1})$.

\subsection{Settings for Dialogue Quality Evaluation }

We evaluated the quality of the synthetic dialogues using the GPT-4 model. The assessment focused on language fluency, logical flow of topics, and the coherence of questions within each session. Conversations were rated on a scale from 1 to 10, where 1 indicated poor quality and 10 represented excellent quality. The overall score was calculated by averaging the ratings across all evaluated conversations. While this approach may exhibit preference bias toward GPT-generated content, it provides consistent comparative assessment across methods. The prompt for this evaluation can be found in Appendix \ref{prompt:session_rating}.

Since our task setting is relatively new, we compared our method against the following baselines:

\begin{enumerate}
    \item \textbf{Random}: Intents and questions were randomly selected without any domain knowledge, although the turn distribution $P(T)$ was maintained.

    \item \textbf{HMM}: We used the classic HMM with random sampling for emissions, as outlined in Section~\ref{ori_hmm}.

    \item \textbf{Golden}: This baseline consisted of actual customer conversations sampled from online chat logs, serving as a reference for natural dialogue quality.
\end{enumerate}

\subsection{Settings for Downstream Task}

To further validate our approach, we test the models on the MTIC task. In this task, given a conversation history of user questions $h_t$ and the last question $q_t$, the goal is to predict the intent $I_t$ of the last question using all previous questions $q_{1...t-1}$ as context. The test sets were drawn from online chat logs, with the intents of the last questions manually annotated. Earlier questions in the conversations were not annotated. We excluded single-turn conversations to focus on the multi-turn aspect. We designed several experiments to understand the impact of our methods.

\subsubsection{Baseline: Single-Turn Data (ST)}

As a starting point, we trained the model using only the single-turn question-intent pairs from our datasets. The model was optimized using standard cross-entropy loss for intent classification.

\subsubsection{Incorporating Multi-Turn Data (MT)}

To evaluate the effect of multi-turn data, we added the generated multi-turn dialogues from the MINT-E corpus to the training set. Each multi-turn conversation was broken down into training samples by concatenating the conversation history up to each turn:
\vspace{-.1cm}
\[
\{ (q_1, I_1), (q_1, q_2, I_2), \dots, (q_1, q_2, \dots, q_t, I_t) \}
\]

This approach allowed the model to learn from the context provided by earlier turns.

\subsubsection{Multi-Task Learning Strategies}

We explored two multi-task learning strategies to enhance the model's performance:

\begin{itemize}
    \item \textbf{Binary Classification for Response Ranking (RR)}: We introduced an auxiliary task where the model predicts whether a given response is of high quality. Responses were labeled as high-quality ($1$) or lower-quality ($0$), and the model was trained using binary cross-entropy loss alongside the intent classification loss.

    \item \textbf{Contrastive Learning for Response Ranking (CR)}: Instead of assigning labels, we used contrastive learning to teach the model to differentiate between better and worse responses. The model learned to assign higher scores to superior responses within a pair.
\end{itemize}

\subsubsection{Training Details}

For both multi-task learning approaches, the intent classification and response ranking tasks shared the same encoder weights (XLM-RoBERTa-base). To balance the number of samples for the response ranking task, we randomly selected responses from different intents to serve as negative examples.

We fine-tuned the XLM-RoBERTa-base model with a learning rate of $2 \times 10^{-5}$, using a batch size of 32 over three epochs. The hyperparameter $\lambda$, which controls the weight of the contrastive loss, was set to 0.3 based on validation results.

\subsection{Results and Discussion}

\subsubsection{Quality of Generated Dialogues}

Table~\ref{table:results_rating} shows the quality ratings for the generated dialogues across different markets. Our generated dataset, \textbf{MINT-E}, consistently outperformed the baselines in every market. Interestingly, our generated dialogues sometimes received higher average ratings than the golden set, as users often provide brief or out-of-context queries, whereas the language model generates more detailed and contextually relevant questions.


To illustrate, here are examples transitioning from the intent "Cancel order" to "Expedite delivery":

\textbf{Example 1 (Human) - Rating: 4}

\begin{quote}
\textbf{$q_1$}: "Cancel my order."\\
\textbf{$q_2$}: "Speed up parcel delivery."
\end{quote}

\textbf{Example 2 (LLM-Generated) - Rating: 8}

\begin{quote}
\textbf{$q_1$}: "Hi, please cancel my order as I needed it by the 3rd for an event."\\
\textbf{$q_2$}: "Is there any way to get it delivered by tomorrow instead of canceling? I really need it."
\end{quote}

The dialogue in the human example lacks coherence, while the LLM-generated example provides context and a logical flow between the two intents.

\begin{table}[t]
\centering
\scalebox{.85}{
\begin{tabular}{|l|c c c c c c c c|c|}
\hline
Model & BR & ID & MY & PH & SG & TH & TW & VN & Avg. \\
\hline
Golden & 7.60 & 6.96 & 6.97 & 7.06 & 7.55 & 7.05 & 7.68 & 7.14 & 7.25 \\
Random & 6.62 & 6.39 & 5.88 & 6.31 & 6.45 & 6.57 & 7.19 & 6.43 & 6.48 \\
Hmm & 7.22 & 6.83 & 6.55 & 6.87 & 6.96 & 7.25 & 7.56 & 6.85 & 7.01 \\ 
MINT-E & \textbf{8.20} & \textbf{7.75} & \textbf{7.85} & \textbf{7.79} & \textbf{7.59} & \textbf{8.12} & \textbf{7.84} & \textbf{7.97} & \textbf{7.89} \\
\hline
\end{tabular}}
\caption{Quality ratings of generated dialogues across different methods and markets. Best performance is highlighted in bold.}
\label{table:results_rating}
\vspace{-.7cm}
\end{table}

\begin{table}[t]
\centering
\scalebox{.83}{
\begin{tabular}{|l|c c c c c|c|c|}
\hline
Setting & ID & MY & PH & SG & TW & Avg & Avg(EN) \\
\hline\hline
ST & 66.67\% & 59.50\% & 50.53\% & 64.86\% & 63.72\% & 61.05\% & 58.30\% \\
MT & 63.11\% & 63.13\% & 51.60\% & 66.06\% & 62.24\% & 61.23\% & 60.27\% \\
ST+RR & 67.70\% & 57.35\% & 47.33\% & 66.67\% & 62.83\% & 60.38\% & 57.12\% \\
ST+CR & 65.73\% & 58.15\% & 49.25\% & 66.37\% & 64.60\% & 60.82\% & 57.92\% \\
MT+RR & 64.23\% & 62.16\% & 52.45\% & 66.06\% & 61.65\% & 61.31\% & 60.23\% \\
MT+CR & 63.67\% & 62.46\% & 52.24\% & 67.72\% & 61.65\% & \textbf{61.55\%} & \textbf{60.81\%} \\
\hline
\end{tabular}}
\caption{Impact of different training configurations on Multi-Turn Intent Classification accuracy. Best performance (averages across all markets and English-only markets) is highlighted in bold.}
\label{table:results_downstream}
\vspace{-.8cm}
\end{table}

\subsubsection{Downstream Task Performance}

The impact of different training configurations on the MTIC task is presented in Table~\ref{table:results_downstream}. Incorporating multi-turn data generally improved performance compared to using only single-turn data. The multi-task learning strategies improved average accuracy by 0.5\% on average across 5 markets, particularly in English-speaking markets.

However, in some non-English markets like Indonesia (ID) and Taiwan (TW), including multi-turn classification data didn't consistently boost performance. This might be due to limitations in the language model's ability to handle specific languages, which affect the quality of generated dialogues, further discussed in section 6.2. Instead, adding only the response ranking data trained with multi-task learning achieves better results on TW and ID. Because we generate 2 responses via 2 different LLM for response evaluation separately, increasing the chance of generating a proper answer. The response ranking as an auxiliary task can also not affect the classification task directly. With appropriate $\lambda$ setting for loss combination, we can avoid Seesaw Effect \cite{Ruder2017AnOO} on multi-task learning and alleviate poor generation quality issues on non-English markets.

\subsubsection{Discussion and Limitations}

Our experiments show that integrating domain knowledge with LLM, as done in MINT-CL, generally enhances dialogue quality and improves performance on downstream tasks. The results suggest that leveraging multi-turn data and multi-task learning is beneficial, especially in markets where the language model performs well.

However, challenges persist in low-resource languages where LLMs are less effective. Future work could focus on improving language models for these languages or exploring alternative methods to generate high-quality dialogues in multilingual settings.
\begin{table}[t]
\centering
\scalebox{.86}{
\begin{tabular}{|l|c c c|c|}
\hline
Setting & BR & TH & VN & Avg \\
\hline\hline
ST & 65.89\% & 65.15\% & 82.24\% & 65.89\%  \\
MT & 61.41\% & 56.22\% & 72.46\% & 61.41\%  \\
\hline
\end{tabular}}
\caption{Single-turn (ST) and multi-turn (MT) results on low-resource languages.}
\label{table: multilingual result}
\vspace{-.9cm}
\end{table}

\begin{table}[t]
\centering
\scalebox{.8}{
\begin{tabular}{|l| c c|} 
 \hline
 Items & MINT-E & Golden \\ [0.5ex] 
 \hline
 \# Sessions & 20k & 2k \\
 \# Questions & 44k & 4k \\
 \# Words & 482k & 29k \\
 \# Avg. \# questions per session & 2.92 & 2.78 \\
 \# Avg. \# words per question & 11.00 & 7.06 \\ [1ex]
 \hline
\end{tabular}}
\caption{Comparison of MINT-E statistics with the golden set sampled from real online chat logs.}
\label{table:statistics_golden}
\vspace{-1cm}
\end{table}

\section{Ablation Study}

\subsection{Component Ablation}

\subsubsection{Effect of Removing LLM Enhancement}
We examine the impact of individual components in the generation pipeline, specifically the HMM base model and the LLM enhancement, on overall performance.

Firstly, we explore using HMM without LLM enhancement by randomly sampling utterances from similar questions under the intents generated by HMM. As shown in Table~\ref{table:results_rating}, the dialogue quality decreases by an average of 0.88 points without LLM generation. This drop occurs because the original HMM is limited by its emission independence, generating utterances conditioned solely on the current intent without considering prior context. Nevertheless, the HMM-only variant still achieves competitive performance compared to the golden baseline, demonstrating the strength of intent modeling.

\subsubsection{Impact of Removing the Chain of Intent}
Next, we investigate the effect of removing the "chain of intent" generated by the HMM, which relies on prior distribution data from real chat logs. When the Markovian constraint is removed and intent is randomly sampled at each turn, we observe a decline in the natural flow of conversation. This results in reduced human-likeness, limited topic diversity, and a lack of authentic multi-turn conversational dynamics. Despite this, high-quality intent sequences remain effective in generating intent-aware dialogues, supporting prior findings that adding intent structure improves dialogue generation quality~\cite{kong2024platolm, Soudani2024ASO}.

\subsection{Generation Quality in Low-Resource Languages}

\subsubsection{Effect of Token Fertility}
Table~\ref{table: multilingual result} illustrates that MINT-CL shows weaker results on classification tasks in low-resource languages where LLM-generated content may be of lower quality. We investigated this issue from two key aspects: token fertility and average words per sentence.

Token fertility measures the number of sub-words generated per word when tokenizing~\cite{Rust2020HowGI}. Higher token fertility typically results in lower generation quality. For instance, \cite{Nguyen2023SeaLLMsL} found that Thai (TH) and Vietnamese (VN) languages require 9X and 4X more tokens, respectively, compared to English when using ChatGPT's tokenizer. This higher tokenization complexity negatively impacts generation quality. Similarly, Brazilian Portuguese (BR) is not encoded efficiently by LLMs without additional training data, as it differs significantly from European Portuguese in terms of vocabulary, grammar, pronunciation, and cultural expressions~\cite{kluge2024teenytinyllama}. These linguistic challenges lead to degraded generation quality and performance drops in low-resource languages.

\subsubsection{Impact of Sentence Length}
Additionally, we observe that the average number of words per question in MINT-E exceeds that of the golden dataset by 4 words, as shown in Table~\ref{table:statistics_golden}. This discrepancy in sentence length may introduce extraneous information during model training, further affecting classification task performance. Overall, these analyses confirm that LLM generation quality plays a crucial role in the effectiveness of our pipeline, particularly in low-resource language scenarios.

\vspace{-.1cm}
\section{Related Work}

\textbf{Multi-turn Dialogue Generation (MTDG)} emerges from both academia and industry due to its relevance to real-world applications, such as chatbots and information-seeking systems, in production~\cite{liu2024balancing,liu2024responsible}. However, they face challenges such as the data sparsity \cite{Zhang2018TailoredST, Zhang2018ReinforcingCF, Li2017AdversarialLF}. Moreover, MTDG requires more complex information and constraints \cite{zhang2020modeling, Chen2018HierarchicalVM, Zhang2018ContextSensitiveGO,mu2023modelling}, posing additional challenges. In general, dialogue generation is categorised into open-domain generation and task-oriented generation \cite{lv-etal-2023-dialogps}. 

\textbf{Open-domain generation} is a context-sensitive process that spans multiple turns, with the model learning to generate appropriate yet open-ended responses based on previous utterances (i.e., the context), which can be categorized into preLLM and postLLM \cite{soudani2023data}. \textbf{PreLLM} methods include the HRED \cite{Serban2015BuildingED} series of models. This hierarchical encoder-decoder framework effectively captures contextual information and incorporates latent variables as intermediate states to generate diverse responses, such as variable HRED \cite{Serban2016AHL}, MrRNN \cite{Serban2016MultiresolutionRN} and DialoGPS\cite{lv-etal-2023-dialogps}. Other related research adopts deep-learning models \cite{santhanam2019survey} enhanced by internal knowledge to ensure on-topic response generation\cite{yu2022survey, madotto2018mem2seq, wang2020improving,mu2024label,mu2022revision}. Popular architectures include recurrent neural networks (RNNs) \cite{li2015diversity}, memory networks \cite{Chen2018HierarchicalVM} and transformers \cite{lin2020hierarchical,olabiyi2023multi}. Those works introduce topic words \cite{xing2017topic},  emotion \cite{song2019generating, zhou2018emotional}, and persona \cite{zhang2018personalizing, zhou2018emotional} into the generation process to help deliver better dialogue responses. Most of the research still adopts supervised models, which require much training data for better performance. 
The advent of LLM has created new possibilities for automated text generation. \textbf{PostLLM} methods leverage powerful LLM to utilize prompt or instruction to generate instruction-response pairs automatically. Due to the excellent generation ability of LLM, most research on generation tasks focuses on developing high-quality instruction for multi-turn generation \cite{wang2022self,ding2023enhancing} and enhancing the multi-turn instruction following ability with human feedback \cite{Sun2023ParrotEM,ouyang2022training}. However, these open-domain generation methods do not explicitly adopt domain-specific intents into dialogue generation, so their work is hardly used in other downstream NLP tasks, such as intent or emotion classification tasks. 

\textbf{Task-oriented generation} is goal-driven and limited to specific domains for task-oriented dialogue (TOD) \cite{soudani2023data}. It has been widely deployed in many industries as it delivers more efficient customer support \cite{moghe2022multi3nlu++,deng2022user}. The popular approach centres around user simulation \cite{lin2021domain, tseng2021transferable}, where two agents interact to process the dialogue history and create a belief state for each side of the conversation, called self-play. They adopt HMM \cite{Cuayhuitl2005HumancomputerDS}, partially observable Markov decision processes (POMDP), as a user model for spoken dialogue generation. Recent work also adopts LLM as a user simulator \cite{kong2024platolm} with reinforcement learning \cite{min_et_al_2023} and in-context learning framework \cite{terragni2023context,Askari2024SelfseedingAM}. This work considers the limited number of dialogue acts, which can't include all types of intents within the chatbot.

\textbf{Multi-Turn Intent Classification} focuses on understanding user intents within the context of ongoing dialogues. \citet{larson2019evaluation} highlighted the challenge of distinguishing out-of-scope queries, emphasizing the need for robust intent classification models. To enhance interpretability in intent identification, \citet{yang2020reinforcement} utilized reinforcement learning over knowledge graphs, demonstrating improved performance in multi-turn settings. \citet{tran2022adaptive} proposed an adaptive global-local context fusion mechanism, effectively integrating dialogue context for enhanced intent classification and slot filling. To address the complexity of multi-turn conversational contexts in the e-commerce domain, \citet{junhua2024lara} developed LARA, a Linguistic-Adaptive Retrieval-Augmentation framework across multiple languages. Furthermore, \citet{fan2022building} developed a system leveraging pre-trained language models for fine-grained query understanding, aligning with our focus on intent-aware dialogues in specific domains. Additionally, \citet{hao2023intentdial} introduced a method employing reinforcement learning to construct an intent graph for identification, highlighting the importance of capturing intent transitions in dialogues.

\section{Conclusion}

Effective Multi-Turn Intent Classification (MTIC) in chatbot systems requires large-scale, domain-specific, multilingual multi-turn dialogue datasets. Creating such datasets require tremendous time and resources. To tackle this challenge, we first proposed \textbf{Chain-of-Intent} mechanism, which leverages LLM-enhanced HMMs to model sequences of user intents based on prior knowledge extracted from historical chat logs. By sampling intent transitions and turn distributions, the LLM effectively capture realistic intent flows and generates coherent and contextually appropriate multi-turn dialogues. 
Furthermore, we introduced \textbf{MINT-CL}, a framework that employs multi-task contrastive learning for MTIC, which is able to learn robust representations by incorporating additional training signals, thereby improving intent classification accuracy in multilingual and resource-constrained settings. Lastly, we released \textbf{MINT-E}, a multilingual, intent-aware, multi-turn e-commerce dialogue corpus generated using our framework. 
Our experiments demonstrate promising results in quality evaluation of the dataset and model performance for MTIC downstream tasks across languages.


{
\vspace{3mm}
\footnotesize
\noindent{\textbf{Acknowledgements}}.
This research is supported in part by the Ministry of Education, Singapore, under its Academic Research Fund Tier 2 (Award No. MOE-T2EP20123-0015). Any opinions, findings and conclusions, or recommendations expressed in this material are those of the authors and do not reflect the views of the Ministry of Education, Singapore.
}



\appendix
\lstnewenvironment{prompt}[1][]{%
  \lstset{
    basicstyle=\ttfamily\scriptsize,
    frame=tb,
    escapeinside=||,
    breaklines=true,
    #1
  }%
}{}

\section{Prompts}
The prompts for GPT-4 will typically involve three distinct roles: system, user, and assistant. In the following examples, the messages for each role are clearly distinguished by their respective role, denoted by \textbf{[{role}]} tags. Variables are encapsulated in between "\{" and "\}" symbols.

\subsection{Prompt for question generation}\label{prompt:question_generation}
\begin{prompt}%[title={Prompt for question generation},mathescape=true]
|\textbf{[system]}|
Forget that you are an OpenAI chatbot assistant, pretend that you are a customer on an online shopping platform, and the user is a customer service chatbot (CS chatbot) to help you with your enquiries. Your task is to complete the conversation with an enquiry with intention {cn}. This is a conversation with a chatbot, so make it informal. Besides, your enquiry should be related and continuous of the previous context only if possible, that is, without deviating from the definition of example enquiries. However, it is important to refer to the example enquiries below, strictly follow their meaning and scope when asking enquiries. The meaning of the inquiry should be consistent with the intention after considering the context. Answer in the language used in your previous enquiry.
You may use coreference to make it more natural, substitute the subject mentioned in previous enquiries with pronouns. For example, you may say "Cancel it" instead of "Cancel the ShopeeFood order" if "ShopeeFood" had been mentioned by you before. Most importantly, DO NOT deviate from the meaning of the example enquiries, follow the definition faithfully, do not add unrelated information.
Lastly, for your attention, the utterance should have intention {intent}. Omit the details included in the previous enquiries and DO NOT start with acknowledgement like "ok", "i see", "one more question", "sorry" etc, keep it simple.
Example enquiries for the intent:
 - {sampled single-turn question 1-3}

Remember, you are the customer, DO NOT act like you are providing help. DO NOT thank the chatbot. DO NOT apologize.

|\textbf{[user]}| {empty first message}
|\textbf{[assistant]}| {$q_1$}
|\textbf{[user]}| {$a_1$}
|\textbf{[assistant]}|{completion starts here}
\end{prompt}

\subsection{Prompt for answer generation}\label{prompt:answer_generation}
\begin{prompt}%[title={Prompt for answer generation},mathescape=true]
|\textbf{[system]}|
Pretend that you are a customer service agent for an e-commerce platform. Rules:
 - You always have the solution, be helpful to the user's enquiries.
 - Make it as short as possible, the answer should be under 20 words.
 - Please answer in the same language as the latest user inquiry.

|\textbf{[user]}|{$q_1$}; |\textbf{[assistant]}|{$a_1$}
|\textbf{[user]}|{$q_t$}; |\textbf{[assistant]}|{completion starts here}
\end{prompt}

\subsection{Prompt for answer rating}\label{prompt:answer_rating}
\begin{prompt}%[title={Prompt for answer rating},mathescape=true]
|\textbf{[system]}|
Please act as an impartial judge and evaluate the quality of the response provided by an AI assistant to the final customer enquiry in the chat history displayed below. Your evaluation should consider factors such as the helpfulness, relevance, accuracy, depth, creativity, and level of detail of the response. Be as objective as possible. Rate the response on a scale of 1 to 10. DO NOT provide any explanation, strictly return only a rating.

|\textbf{[user]}|
#chat history
Customer: {$q_1$}
Assistant: {$a_1$}
Customer: {$q_2$}
# Response
Assistant: {$a_2$}
|\textbf{[assistant]}|
{completion starts here}
\end{prompt}

\subsection{Prompt for session quality rating}\label{prompt:session_rating}
\begin{prompt}%[title={Prompt for session quality rating},mathescape=true]
|\textbf{[system]}|
Please act as an impartial judge and evaluate the quality of the conversation generated artificially by an AI assistant. The conversation generated happens on an e-commerce platform, it includes enquiries from a customer directed to a customer service chatbot (cs chatbot).Your evaluation should consider factors such as fluency of language, topic flow reasonableness, and continuity of messages. Furthermore, The language used shouldn't matter in this test.
Rate the quality on a scale of 1 to 10, you may use this as the standard:
1 > Extremely not fluent and unnatural. There is abrupt change of topic between enquiries with no logical connection to previous dialogues. There is any contradiction in the enquiries.
10 > Extremely fluent and natural.


|\textbf{[user]}|
# Conversation to be evaluated
customer: {$q_1$}
cs chatbot: {$a_1$}
customer: {$q_2$}
cs chatbot: {$a_2$}

STRICTLY answer only with a score, DO NOT explain or elaborate with any words.

|\textbf{[assistant]}|
{completion starts here}
\end{prompt}

\section*{GenAI Usage Disclosure}
We employ generative AI as part of the data augmentation process within our proposed framework, as detailed in earlier sections of the paper. We also occasionally used generative AI tools to assist with grammar, spelling and minor edits on text originally written by the authors.

\bibliographystyle{ACM-Reference-Format}
\bibliography{sigconf}

\end{document}